# A Mobile Application for Flower Recognition System Based on Convolutional Neural Networks

Mustafa YURDAKUL[1*], Enes AYAN[2], Fahrettin HORASAN[3], Şakir TAŞDEMİR[4]

[1,4]Selçuk University, Department Computer Engineering, Konya, Türkiye

[2,3]Kırıkkale University, Department of Computer Engineering, Kırıkkale, Türkiye

*Correspondence: mustafayurdakul@kku.edu.tr

ORCIDs:

Mustafa YURDAKUL: https://orcid.org/0000-0003-0562-4931
Enes AYAN: https://orcid.org/ 0000-0002-5463-8064
Fahrettin HORASAN: https://orcid.org/0000-0003-4554-9083
Şakir TAŞDEMİR: https://orcid.org/0000-0002-2433-246X

**Abstract:**

A convolutional neural network (CNN) is a deep learning algorithm that has been specifically designed for computer vision applications. The CNNs proved successful in handling the increasing amount of data in many computer vision problems, where classical machine learning algorithms were insufficient. Flowers have many uses in our daily lives, from decorating to making medicines to detoxifying the environment. Identifying flower types requires expert knowledge. However, accessing experts at any time and in any location may not always be feasible. In this study a mobile application based on CNNs was developed to recognize different types of flowers to provide non-specialists with quick and easy access to information about flower types. The study employed three distinct CNN models, namely MobileNet, DenseNet121, and Xception, to determine the most suitable model for the mobile application. The classification performances of the models were evaluated by training them with seven different optimization algorithms. The DenseNet-121 architecture, which uses the stochastic gradient descent (SGD) optimization algorithm, was the most successful, achieving 95.84 % accuracy, 96.00% precision, recall, and F1-score. This result shows that CNNs can be used for flower classification in mobile applications.

## 1. INTRODUCTION

Flowers are a type of plant with various species. They play a crucial role in living organisms and the natural world. Flowers are used in various fields in our daily life such as decoration, drug-making[1], and environmental detoxification[2]. They maintain the humidity balance of the environment and can have positive psychological effects through their visual textures and fragrances [3]. According to a study carried out in 2014, employees who worked in an office with 30 different types of flowers for four minutes experienced a relaxing effect on the employees [4]. There are an estimated 400,000 species of flowers worldwide[5]. Although there are thousands of flower species that can be seen in daily life, only a few of them are recognized by humans. Flower classification is a crucial topic in botany, demanding specialized knowledge and expertise. Flower type determination with classical methods is made according to features such as color, pattern, scent, or shape of the flower. However, the classification process using classical methods is inadequate due to the similarities between flowers. The training of an expert in this field is a time-consuming and laborious process. Training an expert in this field is a time-consuming and laborious process. Furthermore, if knowledge and experience is not passed on, it may be lost forever with the death of the expert. For these reasons, the demand for computer-aided classification systems that provide reliable and precise results is growing. Artificial intelligence-based image classification is utilized in various fields, including health [6], education [7], agriculture [8], and defense [9]. CNNs have become a popular research area in computer vision due to their ability to automatically extract the necessary features for image classification from raw image data. They have been utilized to address various agricultural issues, including the classification of crop pests [10], identification of plant species [11], and classification of plant diseases [12]. In the literature, classification of flower species using artificial intelligence is divided into two: studies using classical machine learning and studies using deep learning methods. Some studies with classical machine learning methods are as follows. Mohd-Ekhsan et al. (2014) [13] performed a classification process on images of 18 different flower species with the help of a back-propagation artificial neural network algorithm. In the study, seven different low-level features were extracted for classification. The proposed method achieved 70.19% accuracy. Pardee et al. (2015) [14] used image processing and machine learning algorithms for the classification of 15 different flower species. The GrabCut algorithm was used for the segmentation of the flower from the background, and the color histogram of images was used as a feature for the classification. The random forest algorithm was used to classify images. The algorithm achieved 80.67% accuracy. Lu et al. (2017) [15] extracted features, such as color and wavelet entropy, from 157 petal images. These features were used to train machine learning algorithms, including support vector machines (SVM), kernel-based extreme learning machines, decision tree machines, and weighted K nearest neighbors. The weighted K nearest neighbor algorithm achieved the best classification accuracy, with a score the 99.4%. Pinto, Kelur et al. (2018) [16] were used three different machine learning algorithms (K-nearest neighbor, SVM and, logistic regression) to classify the IRIS dataset. The SVM algorithm achieved the highest accuracy rate of 96%. Classical machine learning algorithms need feature extraction algorithms, such as Histogram of Oriented Gradients (HOG) [17], Scale Invariant Feature Transform (SIFT) [18], and Harris [19]. Classification operations can be carried out using the features obtained from the feature extraction algorithms. However, the extracted features from the data that are very similar to each other are insufficient for the classification process. Therefore, the classification has resulted in lower performance. In addition, machine learning algorithms have proven to be inadequate in dealing with the growing amount of data. There has been a significant increase in studies that utilize the CNNs deep learning algorithm for flower classification in recent years. Some of these studies are as follows. Cengil et al. (2019) conducted a study on the classification of 4000 flower images into five different classes using various CNN architectures, including AlexNet[20], ResNet[21], VGG16[22], GoogleNet[23], and DenseNet201[24]. The VGG16 architecture achieved the highest accuracy rate of 93.52% [25]. Sai, Hitesh et al. (2022) [26] performed flower classification using the ResNet architecture on 4242 flower images, classified into 5 classes. The study reported a classification accuracy rate of 83.79%. Although CNNs have been successful in the detection of flowers, the improvement of their accuracy rate is an active research area. In addition, to the best of our knowledge, there hasn't been any study of a CNN-based mobile application for flower classification, and this study is unique in that respect. In this study, three different scale CNN models, large (Xception)[27], medium (DenseNet-121)[24] and small (MobileNet)[28], trained with transfer learning and fine-tuning methods. Ultimately, the most successful model was selected for use in the mobile application. Main contributions of this study are given following.

• MobileNet, Dense121 and Xception CNN models are trained using transfer learning and fine-tuning to classify 16 different flower species. The models were selected as small, medium and large, respectively, taking into account the number of parameters.
• The CNN models were trained using seven different optimization algorithms, and then their performances compared.
• The optimal freeze ratio in fine tuning for the classification problem was determined by adjusting the freezing ratio of the convolutional layers in models.
• A comparison was performed between the performance of global average pooling and flatten layers at the end of the convolution layers of the CNN models.
• A mobile application was developed for end users using the best-performed CNN model and the application's classification performance was tested on various mobile devices.

The rest of the study is organized as follows. Section 2 describes the details of the dataset, used methods, and evaluation criteria. Section 3 provides the results of the study. Section 4 includes discussion and Section 5 is the conclusion of the study.

## 2.MATERİAL AND METHODS

In this study, a mobile application based on CNN was developed to classify 16 different flower species. Three different sizes of CNN models (large, medium and small) were trained using appropriate transfer learning and fine-tuning methods in order to determine the classifier CNN model for the mobile application. The CNN model that performed the best in the tests was utilised for the mobile application. Figure 1 shows a flowchart summarizing the study.

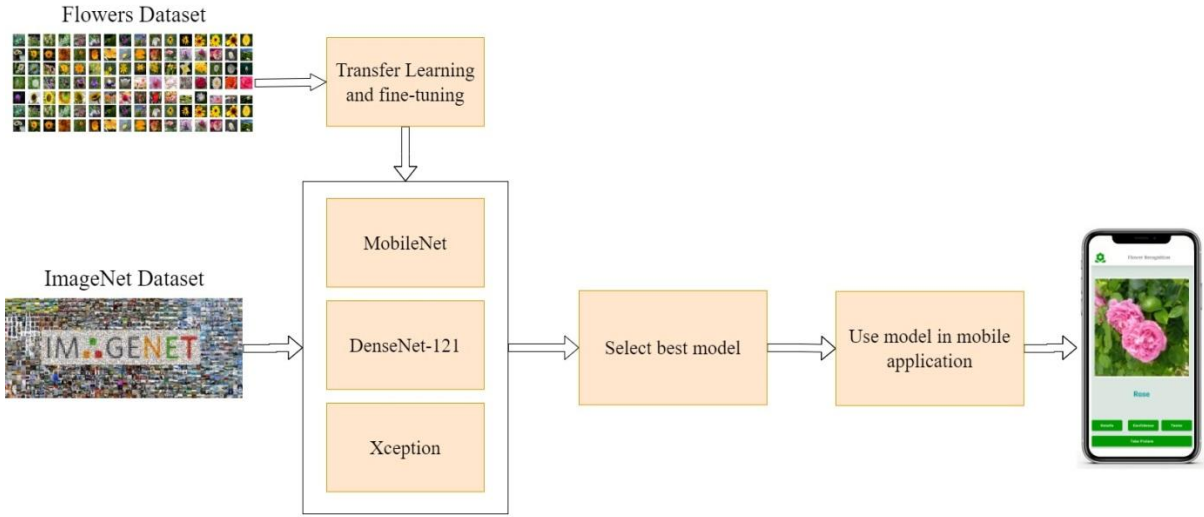

**Fig. 1** Flowchart of proposed study

**2.1 Dataset**

In this study, an open-source dataset was used. It consists of 15,742 RGB images of 16 different flower species[29]. Although the images have different resolutions, they were all scaled to 224x224x3 and 229x229x3. MobileNet and DenseNet models were trained using 224x224x3 scaled images. The Xception model was trained using 229x229x3 scaled images. The data set was divided into three groups: 80% for training, 10% for validation and 10% for testing. Table 1 presents the distribution of flower types in the dataset across the train, validation, and test groups. The dataset appears well balanced, with a similar number of flower species. Fig. 2 shows sample images of the flower types used in the study, along with their corresponding names. The performance of a deep learning model is directly proportional to the size of the training dataset. The training dataset can be augmented if the dataset is insufficient or unstable. Online data augmentation was performed on the training dataset and the validation dataset within the scope of the study. The data augmentation parameters for image manipulation included a rotation range of 0.4, width shift range of 0.2, height shift range of 0.3, shear range of 0.2, and zoom range of 0.2.

**Table 1** Distributions of Dataset

| Flower | Train | Test | Validation | Total |
|---|---|---|---|---|
| Astilbe | 589 | 75 | 73 | 739 |
| Bellflower | 698 | 88 | 87 | 873 |
| Black-eyed Susan | 800 | 100 | 100 | 1000 |
| Calendula | 782 | 99 | 97 | 978 |
| California Poppy | 817 | 103 | 102 | 1022 |
| Carnation | 738 | 93 | 92 | 923 |
| Common Daisy | 784 | 98 | 98 | 980 |
| Coreopsis | 837 | 106 | 104 | 1047 |
| Dandelion | 841 | 106 | 105 | 1052 |
| Daffodil | 776 | 97 | 97 | 970 |
| Iris | 843 | 106 | 105 | 1054 |
| Magnolia | 838 | 106 | 104 | 1048 |
| Rose | 799 | 101 | 99 | 999 |
| Sunflower | 821 | 104 | 102 | 1027 |
| Tulip | 838 | 106 | 104 | 1048 |
| Water Lily | 785 | 99 | 98 | 982 |
| Total | 12586 | 1587 | 1567 | 15742 |

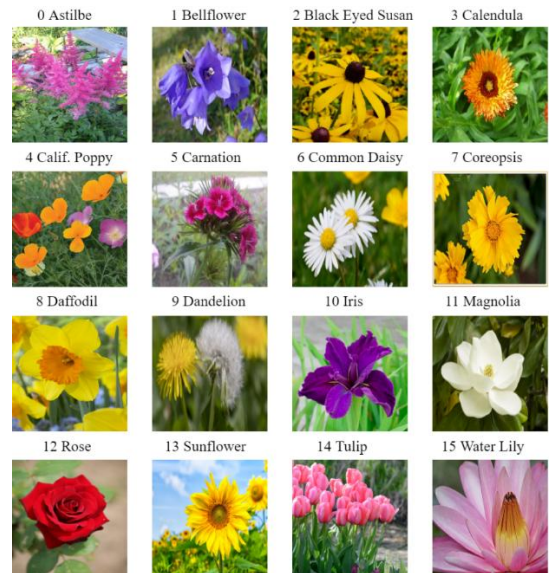

**Fig. 2** Sample Images From the Dataset

## 2.2 Convolutional Neural Network

CNNs are a subfield of deep learning inspired by the visual cortex of the mammalian brain. In recent years, they have successfully solved various computer vision problems, including classification, segmentation, and object detection [30]. Unlike classical machine learning methods, CNNs automatically learn the necessary features to solve the problem using a backpropagation algorithm. A standard CNN architecture consists of three layers: convolutional layers, pooling layers, and fully connected layers. The convolution layer is the main building block of CNN architecture. In the convolutional layers, there are filters with different sizes and numbers to extract low-level and high-level features from the raw image. The pooling layer reduces the size of the input image while preserving important features. Thus, it reduces computational complexity. The final block of CNN consists of fully connected layers, which are traditional artificial neural networks. The features obtained from the stacked convolution and pooling layers are used in the fully connected layers to solve classification or regression problems. In recent years, researchers have developed various CNN architectures to solve computer vision problems. In this study, three different CNN architectures, namely MobileNet, DenseNet121 and Xception, were used for the classification of flower species.

*MobileNet*
Howard et al. (2017) [28] developed MobileNet specifically for use on mobile devices and embedded systems. MobileNet achieves successful classification results with fewer parameters compared to other CNN architectures. The model utilizes the depthwise separable convolutional technique, which enables feature extraction with fewer parameters compared to the standard convolutions used for feature extraction from raw images.

*DenseNet*
Huang et al. (2018) [24] developed the architecture known as DenseNet, named for the dense connections made between layers. The DenseNet architecture utilizes the features of all previous layers as input for each subsequent layer. This results in each layer providing its own features as input for the following layers. The DenseNet architecture has the advantage of enabling feature propagation and reducing the number of parameters by allowing feature reuse. Due to differences in the number of layers, there are four DenseNet architectures: DenseNet-121, DenseNet-169, DenseNet-201, and DenseNet-264. In this study, the DenseNet-121 model was preferred due to the size of the dataset. The architecture of DenseNet-121 includes four dense blocks, three transition layers, and a total of 121 layers.

*Xception*
The Xception architecture was created by Chollet in 2017, drawing inspiration from the InceptionV3. The architecture's most significant innovation is the reverse use of deep separable convolution modules. In this way, it was able to achieve more successful results than Inception V3 on the ImageNet data set. The model comprises 14 modules and a total of 36 convolution layers. All modules, except the first and last, have residual connections.

In the study, only the convolution layers of the MobileNet, DenseNet121, and Xception architectures were utilized, and the fully connected layers were not used. The extracted features have been vectorized by Global Average Pooling (GAP) layer at the end of the convolution layers of all CNN models. After the GAP layer, a dense layer with 16 outputs and softmax activation function is used. CNN architectures are thus structured to compare feature extraction performance of their convolutional layers.

## 2.2 Transfer-Learning Fine-Tuning

Transfer Learning is the use of knowledge and experience gained in solving a problem to solve another similar problem[31]. Due to the nature of convolutional neural networks, basic features such as color, gradient, direction, edge, and corners appear in the first convolution layers, while more specific features of the data emerge towards the final layers. In this study, the weights of a CNN model trained with big data (ImageNet) were used in a network created to solve another problem. The weights were modified directly or depending on the problem. The convolution layers of MobileNet, DenseNet-121, and Xception models pre-trained with the ImageNet dataset were frozen at 0%, 25%, 50%, and 75% ratios and were not included in the training. The main purpose of this freezing process is to determine the fine-tuning rate. Table 3 shows the total number of parameters, the model layers at different freezing ratios and the size of the trained and untrained parameters.

**Table 2** Parameter Sizes of Architectures

| Model | | Total Parameters | Total Trainable Parameters | Total Non-Trainable Parameters | Total Number of Layers | Total Number of Trained Layers |
|---|---|---|---|---|---|---|
| **Xception** | - | 20.894.264 | 20.839.736 | 54.528 | 132 | 132 |
| | %25 | 20.894.264 | 19.734.432 | 1.159.832 | 132 | 33 |
| | %50 | 20.894.264 | 14.891.048 | 6.003.216 | 132 | 66 |
| | %75 | 20.894.264 | 9.511.128 | 11.383.136 | 132 | 99 |
| **DenseNet-121** | - | 7.053.904 | 6.970.256 | 83.648 | 427 | 427 |
| | %25 | 7.053.904 | 6.082.704 | 971.200 | 427 | 106 |
| | %50 | 7.053.904 | 4.648.272 | 2.405.632 | 427 | 213 |
| | %75 | 7.053.904 | 2.072.848 | 4.981.056 | 427 | 320 |
| **MobileNet** | - | 3.245.264 | 3.223.376 | 21.888 | 86 | 86 |
| | %25 | 3.245.264 | 3.192.976 | 52.288 | 86 | 21 |
| | %50 | 3.245.264 | 2.954.256 | 291.008 | 86 | 43 |
| | %75 | 3.245.264 | 2.142.224 | 1.103.040 | 86 | 64 |

The optimizer is an important hyperparameter used in CNN networks. In CNNs, the optimizer plays an important role in finding the appropriate weights that will be minimizing the model error. In this context, each CNN model was trained with 0% freezing with seven different (SGD, Rmsprop, Adagrad, Adadelta, Adam, Nadam, Adamax) optimization algorithms to find the best optimizer. Then models were trained with other freezing ratios (25%, 50%, 75%) using the determined optimizer. The feature maps obtained from stacked convolutional layers must be vectorized so that they can be used in fully connected layers. In the literature, two different methods are used for the vectorization process. The most well-known method is the Flatten method. Each feature map obtained in the Flatten method was converted into a vector. Then these vectors are combined and given as input to fully connected layers. Another method, GAP generates an average value for each feature map in the last convolutional layer. These values are then vectorized and passed to fully connected layers. Fig. 3 provides a visual representation of both methods.

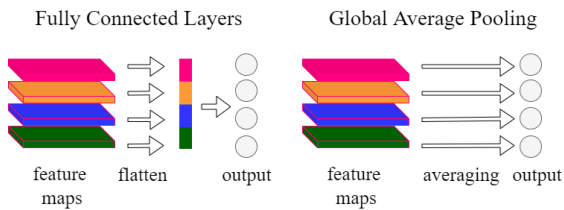

**Fig.3** Fully Connected Layers vs Global Average Pooling

The main advantage of the GAP is the significant reduction in the total number of parameters in the model. In this way, it reduces computational complexity without any performance decrease. In the study, the final convolutional layers of the CNN models were vectorized using GAP and Flatten separately, and the results were then analyzed. Table 2 provide the sizes of the parameters for fine-tuned models based on the usage of GAP. Table 3 provide the sizes of the parameters for fine-tuned models based on the usage of Flatten.

**Table 3** Parameter Counts by Flatten Models

| Model | Total Parameters | Trainable Parameters | Non-Trainable Parameters |
|---|---|---|---|
| MobileNet | 4.031.696 | 4.009.808 | 21.888 |
| DenseNet121 | 7.840.336 | 7.756 | 83.648 |
| Xception | 22.467.128 | 22.412.600 | 54.528 |

### 2.3 Experimental setup

In the study, the Keras deep learning library (Chollet et al., 2015) was used in the implementation and testing stages of the models. All CNN models were trained for 50 epochs, and categorical cross entropy was used as the loss function and the batch size was selected as 32. The training of the models was carried out using Graphics Processing Unit (GPU) provided by Google Colab.

### 2.4 Mobile Application

Mobile devices have become an essential part of our daily lives and mobile applications have made various aspects of our lives easier, from banking and shopping to health and gaming. The objective of this study is to create a mobile application that can accurately and efficiently identify the types of flowers in a person's surroundings. The mobile application that has been developed has the potential to be a fun tool for nature lovers, flower enthusiasts and even children. The most successful CNN model among the trained models was used as the background for the mobile application. The TensorFlow-Lite library was used to transfer this model to the mobile application. A mobile application coded in Java programing language has been developed for the purpose of classifying flowers. The application allows users to take real-time images using their phone camera and identify the type of flower in the image. It can also display the information about the detected flower. Fig. 4 shows the user interface of the developed application. Furthermore, the mobile application was tested with various devices using a test dataset. Additional information can be found in Table 4.

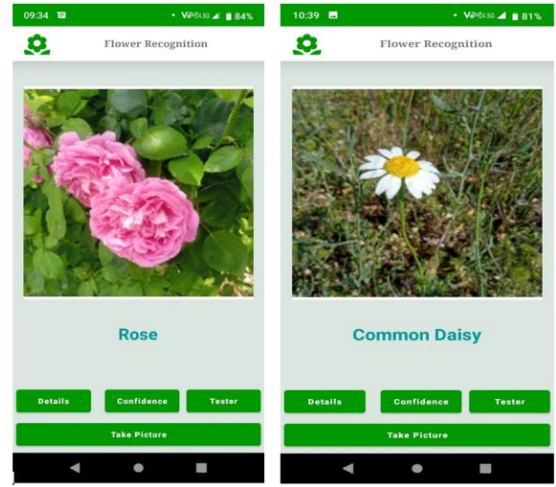

**Fig. 4** The user Interface of Developed Mobile Application

**Table 4** Execute Times by Different Devices

| Device | Specification | Value | Avg. Execute Time (Ms) |
|---|---|---|---|
| **Casper Via S** | **GPU** | PowerVR GE8320, 650 MHz | 279.90 |
|  | **CPU** | 4x 2.0 GHz ARM Cortex-A53 |  |
|  | **OS** | Android 9.0 |  |
|  | **RAM** | 3 GB |  |
|  | **Camera** | 13 MP |  |
| **Xiaomi Redmi Note 9 Pro** | **GPU** | Adreno 618, 750 MHz | 153.56 |
|  | **CPU** | 2.3 GHz ARM Cortex-A76 (Kryo 465) |  |
|  | **OS** | Android 10 |  |
|  | **RAM** | 6 GB |  |
|  | **Camera** | 64 MP |  |
| **Xiaomi Redmi 9** | **GPU** | Mali-G52 MC2, 950 MHz | 68.91 |
|  | **CPU** | 2x 2.0 GHz ARM Cortex-A75 |  |
|  | **OS** | Android 10 |  |
|  | **RAM** | 4 GB |  |
|  | **Camera** | 13 MP |  |
| **Samsung Galaxy A70** | **GPU** | Adreno 612, 845 MHz | 73.47 |
|  | **CPU** | 2x 2.0 GHz ARM Cortex-A76 (Kryo 460) (Gold) |  |
|  | **OS** | Android 10 |  |
|  | **RAM** | 4 GB |  |
|  | **Camera** | 13 MP |  |

**2.5 Performance Metrics**
In the study, test data standalone of the training data was used to evaluate all CNN models. The models' classification performance was compared using objective evaluation criteria. As the study is a multi-classification problem, the metrics were investigated on the basis of both individual classes and class averages. Table 6 shows the equations for the calculation of the criteria and an explanation of their meanings. In addition, the models' predictions are evaluated using confusion matrices. Fig. 5 shows the confusion matrices used for multi-classification operations.

|  | Predicted Class | | | |
|---|---|---|---|---|
|  | 1 | 2 | … | N |
| Actual Class 1 | T1 | F12 | … | F1N |
| 2 | F21 | T2 | … | F2N |
| … | … | … | … | … |
| N | FN1 | FN2 | … | TN |

**Fig. 5** Confusion Matrices for the Multi-classifications

True Positive (TP) in the confusion matrix; the sample is predicted as positive, the sample is actually positive, False Positive (FP); sample predicted positive but actually negative, True (TN); the sample is predicted as negative and is actually negative, False Negative (FN); the sample is predicted as negative and is actually positive.

## 3. RESULTS

This study employed transfer learning and fine-tuning methods to train Xception, MobileNet, and DenseNet-121 architectures for the detection of 16 different flower species. The goal is to select the best optimization algorithm and determine the number of layers to freeze during the transfer learning and fine-tuning process. The performance of the network is directly affected by the optimization algorithm used during the training phase. Therefore, the architectures were trained with different optimizers by a 0% freeze ratio. The classification performances of the CNN models according to the optimizers are listed in Table 6. Once the most effective optimizer had been identified, the classification performance was evaluated by freezing the layers of each model at a different rate. Table 7 presents the test results of the models. Additionally, Fig. 6 illustrates the confusion matrices achieved by the most successful model of each architecture. After determining the best optimizer and freezing rate in the study, the models' performance was measured using the Flatten layer instead of GAP. The results obtained using the Flatten layer are presented in Table 8.

**Table 5** Performance Metrics Formulations

| Metric | Formula |  |
|---|---|---|
| Accuracy | $\sum_{i=1}^{l} \frac{TP_i + TN_i}{TP_i + FN_i + FP_i + TN_i} / l$ | (1) |
| Specificity | $\sum_{i=1}^{l} \frac{TN_i}{FP_i + TN_i} / l$ | (2) |
| Precision | $\sum_{i=1}^{l} \frac{TP_i}{TP_i + FP_i} / l$ | (3) |
| Recall | $\sum_{i=1}^{l} \frac{TP_i}{TP_i + FN_i} / l$ | (4) |
| Error Rate | $\sum_{i=1}^{l} \frac{FP_i + FN_i}{TP_i + FN_i + FP + TN_i} / l$ | (5) |
| F1-Score | $\frac{2x(Avg.Precision \times Avg.Sensivity)}{Avg.Precision + Avg.Sensivity}$ | (6) |

**Table 6** Test Accuracies by Architectures of Model by %0 freeze ratio.

| Model |  | Adam | Adagrad | Adadelta | Nadam | Rmsprop | Adamax | SGD |
|---|---|---|---|---|---|---|---|---|
| **Xception** | Acc. | 0.9124 | 0.9464 | 0.8935 | 0.7208 | 0.8620 | 0.9502 | **0.9533** |
|  | Loss | 0.4716 | 0.1919 | 0.3677 | 2.4394 | 2.3006 | 0.2678 | 0.1978 |
|  | Precision | 0.9300 | 0.9500 | 0.8900 | 0.9100 | 0.8800 | 0.9500 | 0.9500 |
|  | Recall | 0.9100 | 0.9500 | 0.8900 | 0.7200 | 0.8600 | 0.9500 | 0.9500 |
|  | F1-Score | 0.9100 | 0.9400 | 0.8900 | 0.7600 | 0.8600 | 0.9500 | 0.9500 |
| **MobileNet** | Acc. | 0.9338 | 0.9540 | 0.8972 | 0.9287 | 0.9287 | **0.9558** | 0.9546 |
|  | Loss | 0.3830 | 0.1696 | 0.3169 | 0.4152 | 0.4901 | 0.2928 | **0.1902** |
|  | Precision | 0.9400 | 0.9500 | 0.9000 | 0.9400 | 0.9300 | 0.9600 | 0.9500 |
|  | Recall | 0.9300 | 0.9500 | 0.9000 | 0.9300 | 0.9300 | 0.9600 | 0.9500 |
|  | F1-Score | 0.9300 | 0.9500 | 0.9000 | 0.9300 | 0.9300 | 0.9600 | 0.9500 |
| **DenseNet-121** | Acc. | 0.9224 | 0.9502 | 0.9206 | 0.9061 | 0.9155 | 0.9174 | **0.9584** |
|  | Loss | 0.4059 | 0.1748 | 0.2404 | 0.5138 | 0.5078 | 0.5261 | **0.1833** |
|  | Precision | 0.9300 | 0.9500 | 0.9300 | 0.9200 | 0.9200 | 0.9400 | **0.9600** |
|  | Recall | 0.9200 | 0.9500 | 0.9200 | 0.9100 | 0.9200 | 0.9200 | **0.9600** |
|  | F1-Score | 0.9200 | 0.9500 | 0.9200 | 0.9100 | 0.9200 | 0.9200 | **0.9600** |

Table 7 Test Accuracies by Freeze Ratio

| Model | Percentage | Acc. | Loss. | Precision | Recall | F1-Score |
|---|---|---|---|---|---|---|
| **Xception** | **%25** | 0.9458 | 0.2194 | 0.9500 | 0.9500 | 0.9500 |
| | **%50** | 0.9426 | 0.2299 | 0.9400 | 0.9400 | 0.9400 |
| | **%75** | **0.9502** | 0.2039 | 0.9500 | 0.9500 | 0.9500 |
| **MobileNet** | **%25** | 0.9464 | 0.3424 | 0.9500 | 0.9500 | 0.9500 |
| | **%50** | 0.9420 | 0.3909 | 0.9400 | 0.9400 | 0.9400 |
| | **%75** | **0.9514** | 0.3334 | 0.9500 | 0.9500 | 0.9500 |
| **DenseNet-121** | **%25** | **0.9552** | 0.2229 | 0.9600 | 0.9600 | 0.9600 |
| | **%50** | 0.9459 | 0.3180 | 0.9500 | 0.9500 | 0.9500 |
| | **%75** | 0.9413 | 0.3195 | 0.9400 | 0.9400 | 0.9400 |

Table 8 Flatten Layer Resulsts

| Architecture | Acc. | Loss. | Precision | Recall | F1-Score |
|---|---|---|---|---|---|
| MobileNet | 0.9502 | 0.3987 | 0.9500 | 0.9500 | 0.9500 |
| DenseNet-121 | 0.7693 | 0.8067 | 0.7900 | 0.7700 | 0.7700 |
| Xception | 0.9483 | 0.2904 | 0.9500 | 0.9500 | 0.9500 |

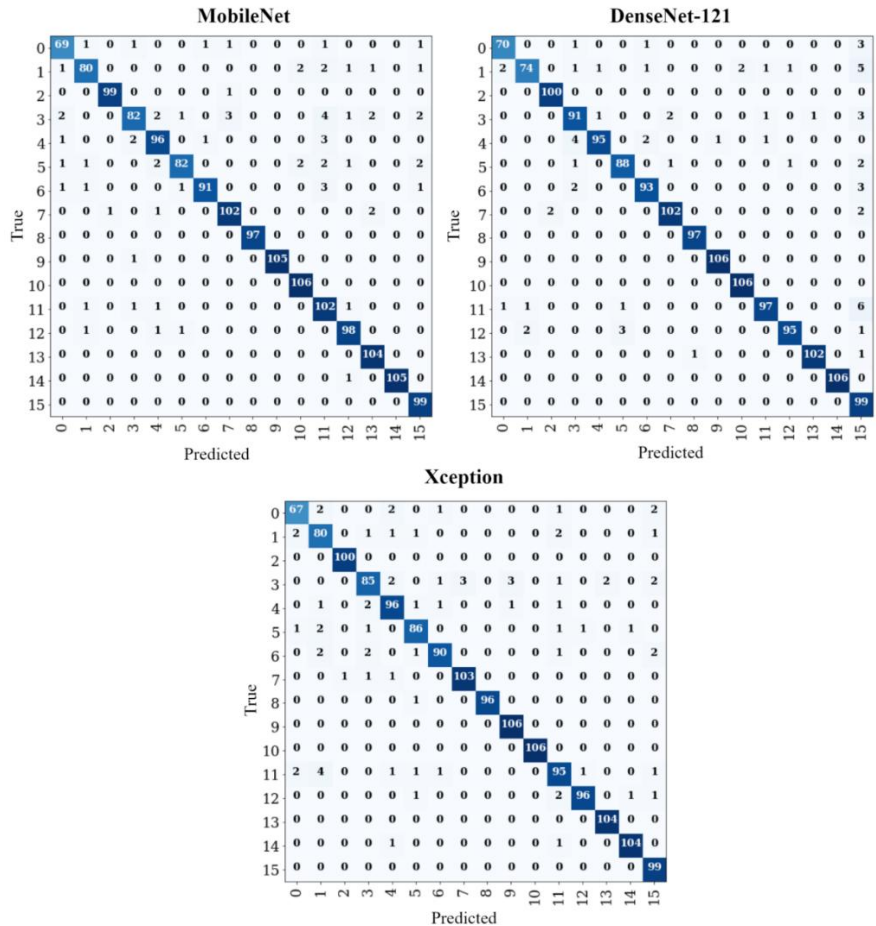

Fig. 6 Confusion Matrices of Best Models

## 4. DISCUSSION

In this study, the performance of CNN models (Xception, DenseNet-121, MobileNet) for flower classification is compared at three different scales using transfer learning and fine tuning methods. During the training of the models, the effect of optimizers and different freezing rates on the classification performance was examined. For this purpose, all models were trained with 7 different optimizers without any freezing. According to the test results, SGD was the best optimizer for Xcepiton and DenseNet-121, while Adamax achieved the best results for MobileNet. After choosing the best optimizer, the models were retrained using the best optimizer by freezing the sequence 25% 50% 75%. The main purpose here is to determine the freezing ratio for this dataset. Table 6 shows that 0% freezing produced better results for all models. In essence, it is acknowledged that retraining the entire model leads to superior performance. The most important factor in this result is the low similarity rate of the dataset we used with ImageNet data. In the study, optimizer and freezing parameters were adjusted using GAP. Then, with these parameters, the models were retrained using Flatten instead of GAP. According to the test results listed in Table 6,7 and 8, it is seen that finishing the convolution layers of the models with GAP gives better results for all models than finishing with Flatten. In addition, when Table 2 and 3 are examined, the parameter decrease in the models when GAP is used can be seen as another advantage of using GAP. A comparison of the studies in the literature and the results we have obtained is given in Table 9. It is understood from this table that our study achieved better results than other studies in terms of flower type and total number of data.

Table 9 Accuracy comparision by literature

| Reference | Flower Class | Total Count | Accuracy |
|---|---|---|---|
| [32] | 5 | 4000 | 93.52 |
| [25] | 5 | 4000 | 93.52 |
| [33] | 102 | 8189 | 87.60 |
| [34] | 8 | 6400 | 85.00 |
| [35] | 5 | 4323 | 91.00 |
| [26] | 5 | 4242 | 83.79 |
| MobileNet | 16 | 15742 | 95.58 |
| DenseNet-121 | 16 | 15742 | 95.84 |
| Xception | 16 | 15742 | 95.33 |

Fig. 7 presents correct and incorrect predictions made by the best performing CNN model, DenseNet-121. The Fig. 7 shows a significant similarity between the misclassified flowers and the actual class.This problem could be seen as one of the limitation of the study and it couldebe eliminated by using more data belonging to the relevant classes.

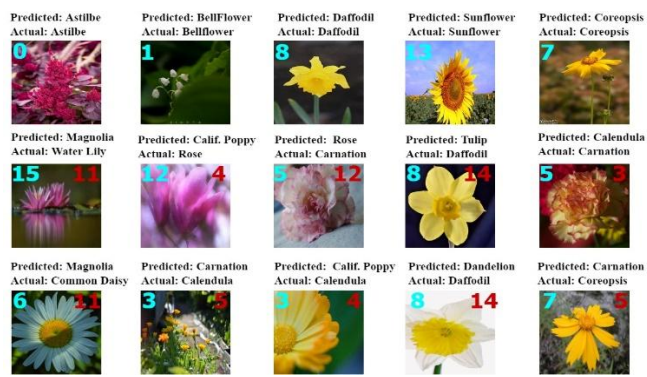

Fig. 7 Some correct and incorrect predictions made by DenseNet-121

## 5. CONCLUSION

In this study, three different sizes of CNN models, namely MobileNet, DenseNet-121 and Xception, were trained using different transfer learning strategies for mobile applications. The DenseNet-121 architecture, which uses the SGD optimization algorithm and GAP, achieved an accuracy rate of 95.84%. Based on this success, a mobile application was developed. Future studies aim to increase the number of classes by expanding the dataset.

**Funding** Not applicable
**Author Contributions** All of the authors have been involved in the discussion, construction,devopment and writing of the paper together.
**Data availability** The dataset used for implementation is a benchmark dataset and its available for free access.
**Conflict of Interests** The authors declare that they have no conflict of interest.
**Ethical Approval** This article does not contain any studies with human participants or animals performed by any of the authors.

## REFERENCES

1. Jafri, M., et al., *Effect of Punica granatum Linn.(flowers) on blood glucose level in normal and alloxan-induced diabetic rats.* Journal of ethnopharmacology, 2000. **70**(3): p. 309-314.
2. Chatterjee, A., et al., *Flower-like BiOI microspheres decorated with plasmonic gold nanoparticles for dual detoxification of organic and inorganic water pollutants.* ACS Applied Nano Materials, 2020. **3**(3): p. 2733-2744.
3. Jo, H.-J. and G.-S. Hong, *Visio-psychological effect of spring flowers blossoms on university students.* Journal of Environmental Science International, 2016. **25**(8): p. 1097-1105.

4. Ikei, H., et al., *The physiological and psychological relaxing effects of viewing rose flowers in office workers.* Journal of physiological anthropology, 2014. **33**(1): p. 1-5.
5. Joly, A., et al. *Lifeclef 2017 lab overview: multimedia species identification challenges.* in *International conference of the cross-language evaluation forum for European languages.* 2017. Springer.
6. Ayan, E., B. Karabulut, and H.M. Ünver, *Diagnosis of pediatric pneumonia with ensemble of deep convolutional neural networks in chest x-ray images.* Arabian Journal for Science and Engineering, 2022. **47**(2): p. 2123-2139.
7. Bebis, G., D. Egbert, and M. Shah, *Review of computer vision education.* IEEE Transactions on Education, 2003. **46**(1): p. 2-21.
8. Kamilaris, A. and F.X. Prenafeta-Boldú, *Deep learning in agriculture: A survey.* Computers and electronics in agriculture, 2018. **147**: p. 70-90.
9. Xin, Y., et al., *Machine learning and deep learning methods for cybersecurity.* Ieee access, 2018. **6**: p. 35365-35381.
10. Ayan, E., H. Erbay, and F. Varçın, *Crop pest classification with a genetic algorithm-based weighted ensemble of deep convolutional neural networks.* Computers and Electronics in Agriculture, 2020. **179**: p. 105809.
11. Yalcin, H. and S. Razavi. *Plant classification using convolutional neural networks.* in *2016 Fifth International Conference on Agro-Geoinformatics (Agro-Geoinformatics).* 2016. IEEE.
12. Sardogan, M., A. Tuncer, and Y. Ozen. *Plant leaf disease detection and classification based on CNN with LVQ algorithm.* in *2018 3rd International Conference on Computer Science and Engineering (UBMK).* 2018. IEEE.
13. Mohd-Ekhsan, H., et al. *Classification of Flower Images Based on Colour and Texture Features Using Neural Network.* in *2010 International Conference on Intelligent Network and Computing (ICINC 2010).* 2014.
14. Pardee, W., P. Yusungnern, and P. Sripian. *Flower identification system by image processing.* in *3rd International Conference on Creative Technology CRETECH.* 2015.
15. Lu, S., et al. *Flower classification based on single petal image and machine learning methods.* in *2017 13th International Conference on Natural Computation, Fuzzy Systems and Knowledge Discovery (ICNC-FSKD).* 2017. IEEE.
16. Pinto, J.P., S. Kelur, and J. Shetty. *Iris flower species identification using machine learning approach.* in *2018 4th International Conference for Convergence in Technology (I2CT).* 2018. IEEE.
17. Dalal, N. and B. Triggs. *Histograms of oriented gradients for human detection.* in *2005 IEEE computer society conference on computer vision and pattern recognition (CVPR'05).* 2005. Ieee.
18. Lowe, D.G. *Object recognition from local scale-invariant features.* in *Proceedings of the seventh IEEE international conference on computer vision.* 1999. Ieee.
19. Harris, C. and M. Stephens. *A combined corner and edge detector.* in *Alvey vision conference.* 1988. Citeseer.
20. Russakovsky, O., et al., *Imagenet large scale visual recognition challenge.* International journal of computer vision, 2015. **115**(3): p. 211-252.
21. He, K., et al. *Deep residual learning for image recognition.* in *Proceedings of the IEEE conference on computer vision and pattern recognition.* 2016.
22. Simonyan, K. and A. Zisserman, *Very deep convolutional networks for large-scale image recognition.* arXiv preprint arXiv:1409.1556, 2014.
23. Szegedy, C., et al. *Going deeper with convolutions.* in *Proceedings of the IEEE conference on computer vision and pattern recognition.* 2015.
24. Huang, G., et al. *Densely connected convolutional networks.* in *Proceedings of the IEEE conference on computer vision and pattern recognition.* 2017.
25. Cengıl, E. and A. Çinar. *Multiple classification of flower images using transfer learning.* in *2019 International Artificial Intelligence and Data Processing Symposium (IDAP).* 2019. IEEE.
26. Sai, A.V., et al. *Flower Identification and Classification applying CNN through Deep Learning Methodologies.* in *2022 International Mobile and Embedded Technology Conference (MECON).* 2022. IEEE.
27. Chollet, F. *Xception: Deep learning with depthwise separable convolutions.* in *Proceedings of the IEEE conference on computer vision and pattern recognition.* 2017.
28. Howard, A.G., et al., *Mobilenets: Efficient convolutional neural networks for mobile vision applications.* arXiv preprint arXiv:1704.04861, 2017.

29. Kaggle. *Flowers Dataset*. 2022 [cited 2022 15.08.2022]; Available from: https://www.kaggle.com/datasets/l3llff/flowers.
30. Alzubaidi, L., et al., *Review of deep learning: Concepts, CNN architectures, challenges, applications, future directions.* Journal of big Data, 2021. **8**(1): p. 1-74.
31. Torrey, L. and J. Shavlik, *Transfer learning*, in *Handbook of research on machine learning applications and trends: algorithms, methods, and techniques*. 2010, IGI global. p. 242-264.
32. Chen, C., et al. *Classification of blurred flowers using convolutional neural networks*. in *Proceedings of the 2019 3rd International Conference on Deep Learning Technologies*. 2019.
33. Qin, M., Y. Xi, and F. Jiang. *A new improved convolutional neural network flower image recognition model*. in *2019 IEEE Symposium Series on Computational Intelligence (SSCI)*. 2019. IEEE.
34. Islam, S., M.F.A. Foysal, and N. Jahan. *A Computer Vision Approach to Classify Local Flower using Convolutional Neural Network*. in *2020 4th International Conference on Intelligent Computing and Control Systems (ICICCS)*. 2020. IEEE.
35. Narvekar, C. and M. Rao. *Flower classification using CNN and transfer learning in CNN-Agriculture Perspective*. in *2020 3rd International Conference on Intelligent Sustainable Systems (ICISS)*. 2020. IEEE.